\documentclass[letterpaper, 10 pt, conference]{ieeeconf}  

\usepackage{amsmath}
\usepackage{amssymb}
\usepackage{mathtools}
\usepackage{graphicx}
\usepackage{microtype}
\usepackage{diagbox}
\usepackage{bm}
\usepackage{booktabs} 
\usepackage{multirow}
\usepackage{float}
\usepackage{booktabs}
\usepackage{hyperref}       
\usepackage{url}            
\usepackage{algorithm} 
\usepackage{algorithmicx}
\usepackage{algpseudocode}
\usepackage{graphicx}
\usepackage{subcaption}
\usepackage{wrapfig}
\IEEEoverridecommandlockouts                              

\title{\LARGE \bf
Novelty-based Sample Reuse for Continuous Robotics Control
}

\author{Ke Duan$^{*,1}$ Kai Yang$^{*,1}$ Houde Liu$^{\dagger,1}$ Xueqian Wang$^{\dagger,1}$
\thanks{*Equal contribution}
\thanks{$^\dagger$Corresponding author}
\thanks{$^1$Shenzhen International Graduate School, Tsinghua University}
}
\begin{document}

\maketitle

\begin{abstract}
In reinforcement learning, agents collect state information and rewards through environmental interactions, essential for policy refinement. This process is notably time-consuming, especially in complex robotic simulations and real-world applications. Traditional algorithms usually re-engage with the environment after processing a single batch of samples, thereby failing to fully capitalize on historical data. However, frequently observed states, with reliable value estimates, require minimal updates; in contrast, rare observed states necessitate more intensive updates for achieving accurate value estimations. To address uneven sample utilization, we propose Novelty-guided Sample Reuse (NSR). NSR provides extra updates for infrequent, novel states and skips additional updates for frequent states, maximizing sample use before interacting with the environment again. Our experiments show that NSR improves the convergence rate and success rate of algorithms without significantly increasing time consumption. Our code is publicly available at \href{https://github.com/ppksigs/NSR_DDPG_HER_for_manipulation}{https://github.com/ppksigs/NSR-DDPG-HER}
\end{abstract}
\section{Introduction}

In reinforcement learning (RL), balancing exploration and exploitation is essential. Exploration encourages agents to gather diverse data by probing uncharted areas of the environment, which helps prevent local convergence and aids in finding global optimal solutions \cite{ladosz2022exploration}. Successful applications of exploration have been demonstrated in Atari games \cite{burda2018exploration,yang2024exploration,ostrovski2017count} and robotic simulations with sparse rewards \cite{ecoffet2021first,pathak2017curiosity}. Conversely, exploitation emphasizes the efficient use of existing samples to enhance the agent's learning process \cite{sutton2018reinforcement}. A key challenge in exploitation is the lengthy interaction time needed for agents to learn effectively.

Extended interaction time arises from the need for the environment to provide the agent with current states and observations, for the agent to generate actions based on its policy, and for the environment to use its internal dynamics model to infer the next state. In simulation environments, this can be computationally intensive \cite{duan2016benchmarking,zhu2017target,levine2016end}. In real-world scenarios, such as with robots, the time required increases as it includes waiting for the physical execution of actions before proceeding \cite{gu2017deep,kober2013reinforcement}. Traditional RL methods typically update their policies once before interacting with the environment again to gather new data. This approach is often time-consuming and inefficient, as it does not fully leverage the historical sample data from previous interactions.

Online RL involves continual interactions with the environment, encompassing both on-policy and off-policy methods. On-policy methods, such as Proximal Policy Optimization (PPO) \cite{schulman2017proximal}, update the policy based on actions taken by the current policy, directly using experiences from the most recent interactions. Off-policy methods, like Deep Q-Networks (DQN) \cite{mnih2013playing}, Soft Actor-Critic (SAC) \cite{haarnoja2018soft}, and Deep Deterministic Policy Gradient (DDPG) \cite{lillicrap2015continuous}, allow the use of past experiences stored in a replay buffer, making them more sample-efficient as they can learn from a larger and more diverse set of experiences. However, despite efficiency, most off-policy deep RL algorithms, especially in continuous control domains, still require numerous interactions to learn meaningful policies. This creates a significant barrier to the application of reinforcement learning algorithms in real-world scenarios, such as robotics, where interactions are costly and time-consuming.

\begin{figure*}[t]
    \centering
    \includegraphics[width=0.85\linewidth]{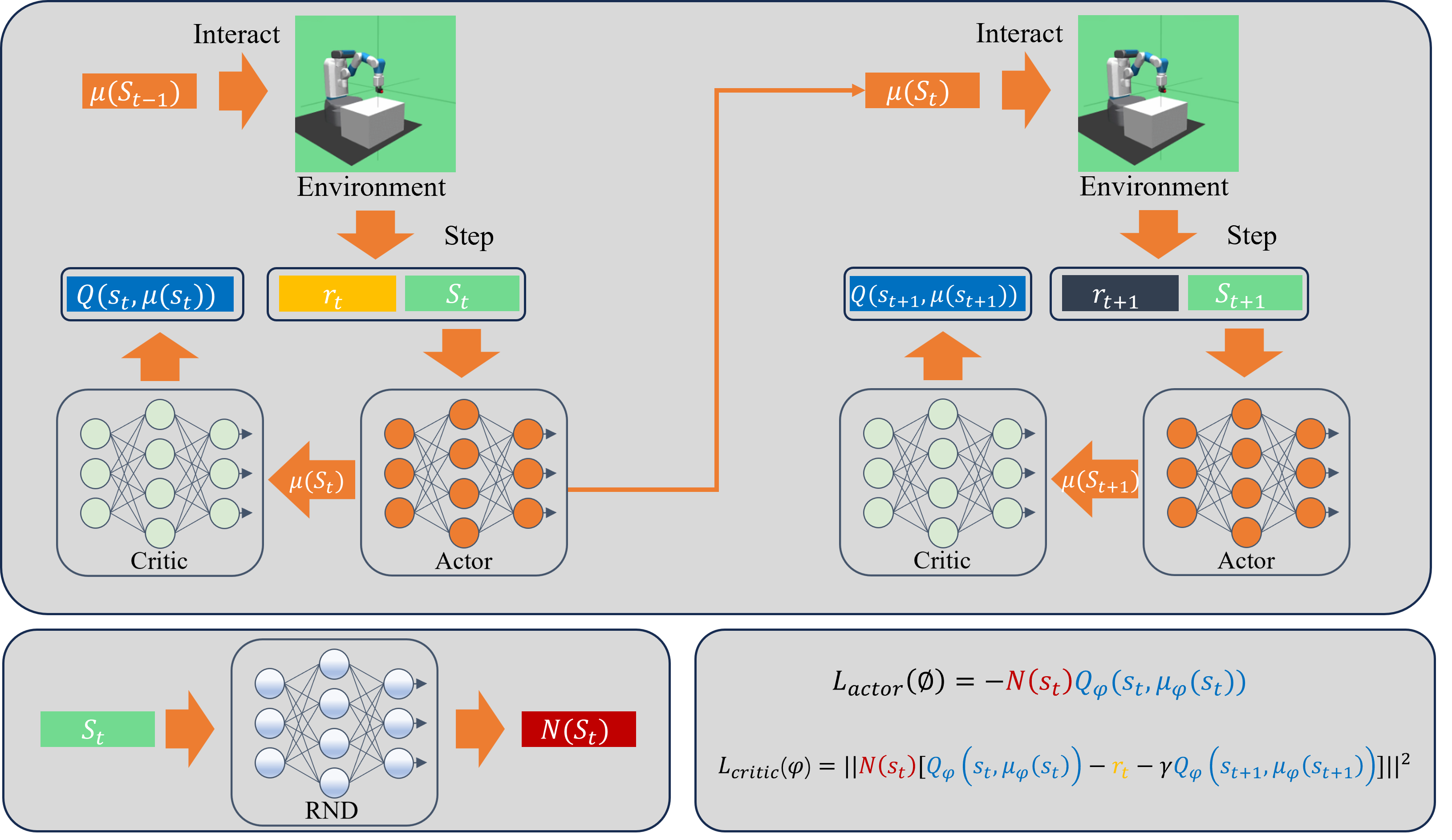}
    \caption{The overview of NSR.}
    \label{overview}
\end{figure*}

Many methods have been developed to reduce overestimation bias in value estimates \cite{2018Addressing,2020Controlling}, often utilizing a high update-to-data ratio \cite{2020Controlling,2022The}. However, these approaches can be complex, incorporating ensembles that require additional neural networks, which increases parameters and demands dynamic models for state transitions, thereby consuming more computational and storage resources. In contrast, our approach simplifies this by repeatedly updating network parameters with data from each environmental interaction. This method, shown to be effective both experimentally and theoretically \cite{lyu2024off}, facilitates extensive sample reuse and repeated updates without the need for constant environmental interactions, thereby enhancing learning efficiency and performance.

Building on this foundation, we develop an advanced method for determining the update frequency of each state based on its novelty. The Random Network Distillation (RND) method \cite{burda2018exploration} measures state novelty through the prediction error of a randomly initialized neural network. High prediction errors indicate that a state is novel and not fully explored, while frequently encountered states exhibit lower prediction errors, reflecting the agent's better understanding of them.

The core idea of our method is that frequently encountered states stabilize their value estimations quickly, requiring fewer updates, while rarely encountered states demand more updates for accurate estimation. By employing the RND method to evaluate state novelty, we can allocate a more appropriate number of updates for each state. This targeted updating mechanism enables the agent to concentrate its learning on states with the highest uncertainty in value estimation, thereby streamlining the learning process and significantly improving performance with fewer environmental interactions.

In summary, we present Novelty-guided Sample Reuse (NSR), a method that allocates updates to samples based on state novelty. Integrated with DDPG, NSR significantly enhances task success rates and manipulator performance in continuous control simulations, requiring minimal additional computation time.

\section{Related Work}
\textbf{Sample-efficient Methods} 
Achieving sample efficiency in reinforcement learning (RL) is essential for reducing the computational cost and time required for agents to learn effective policies. Various methods have been developed to improve sample efficiency, aiming to maximize learning performance with limited interaction data.

One effective approach to sample-efficient RL is model-based reinforcement learning (MBRL). Unlike model-free methods, MBRL algorithms learn an explicit model of the environment dynamics and use it to plan actions \cite{clavera2018model,kaiser2019model,moerland2023model}. By simulating trajectories with the learned model, MBRL agents can explore different action sequences more efficiently and learn from simulated experiences, reducing the need for real-world interactions.

Notable MBRL techniques include Model-Based Policy Optimization (MBPO) \cite{janner2019trust}, which optimizes policies through a combination of model-based planning and policy gradient methods. Another approach is MOPO \cite{yu2020mopo}, which incorporates uncertainty estimation into the model-based planning process to improve robustness to model errors. Probabilistic Ensembles with Trajectory Sampling \cite{chua2018deep} leverages an ensemble of probabilistic models to capture uncertainty in the environment dynamics, enabling better handling of complex, stochastic environments.

Enhancing sample efficiency can be achieved by leveraging prior knowledge or structure in the learning process. Techniques such as transfer learning and meta-reinforcement learning exploit domain-specific insights or pre-trained models to expedite learning. Transfer learning applies knowledge from a source task to a target task, reducing sample requirements by utilizing experience from the source task. Methods like actor-critic with transfer learning extend this to similar environments \cite{parisotto2015actor,rusu2016progressive}. Meta-reinforcement learning enables agents to adapt swiftly to new tasks by learning how to learn from past experiences. Algorithms such as Model-Agnostic Meta-Learning (MAML) facilitate rapid adaptation with minimal data \cite{finn2017model}, proving particularly effective in continuous control tasks \cite{rakelly2019efficient}.

\textbf{Exploration Methods}
Exploration is a fundamental aspect of reinforcement learning (RL) as it enables agents to discover optimal strategies in unfamiliar environments. Various exploration strategies have been developed to address this challenge, aiming to strike a balance between exploiting known information and exploring new possibilities.

Count-based methods provide a systematic approach to exploration by incentivizing visits to less-explored regions of the state space. These methods maintain counts of state-action pairs visited and prioritize exploration in less-frequented areas. For instance, Count-Based Exploration with Neural Density Models (CBE) \cite{ostrovski2017count} utilizes a neural network to estimate the density of states and actions visited.

Curiosity-driven exploration methods harness intrinsic motivation, encouraging agents to explore novel states based on prediction errors or novelty metrics. These methods Exploration by Self-supervised Prediction rewards agents for reducing prediction errors, motivating exploration in unfamiliar regions. The Intrinsic Curiosity Module (ICM) similarly promotes exploration by targeting states with high prediction errors \cite{pathak2017curiosity}. Unlike Curiosity, ICM uses a separate neural network to predict the next state by learning a dynamic model and rewarding exploration based on prediction errors.

Random Network Distillation (RND) is another exploration strategy that encourages agents to explore novel states by measuring the prediction error of a randomly initialized neural network \cite{burda2018exploration}. DRND method extends RND by employing multiple target networks and utilizes statistical constructs to implicitly count states, thereby providing additional rewards to states with fewer visits \cite{yang2024exploration}.

Additionally, techniques such as Upper Confidence Bound (UCB) methods \cite{azar2017minimax} and Thompson Sampling \cite{russo2018tutorial} provide principled frameworks for exploration, enabling agents to balance between exploiting known information and exploring uncertain options. UCB methods maintain estimates of the value of each action and use confidence bounds to balance between exploration and exploitation. Thompson Sampling, on the other hand, maintains a probability distribution over the possible values of each action and samples from this distribution to decide which action to take. These methods ensure that the agent explores different options while also exploiting the best-known strategies.

\section{Preliminaries}

\textbf{MDP.} Our framework is rooted in the MDP formulation outlined in \cite{sutton1998introduction}. Within this framework, an agent interacts with an environment by perceiving a state $s\in \mathcal{S}$ and selecting an action from the set $\mathcal{A}$, representing the state and action spaces, respectively. The transition probability function, denoted by $P(s'|s, a)$, governs the dynamics of transitioning from the current state $s$ to the next state $s'$ upon executing action $a$. Alongside these transitions, the agent receives a scalar reward $r$, determined by the reward function $r:\mathcal{S}\times\mathcal{A}\to\mathbb{R}$, encapsulating the immediate feedback following each action. The overarching goal of the agent is to determine a policy $\pi(a|s)$ that maximizes the cumulative returns of trajectories $\tau = (s_0,a_0,s_1,a_1,...,s_{T-1},a_{T-1})$. This objective is formalized as $\mathcal{J}(\pi) = \mathbb{E}_{\tau}[\sum_{t=0}^{T-1} r\left(s_{t}, a_{t}\right)]$, where $\mathcal{J}(\pi)$ represents the expected cumulative reward over a trajectory under policy $\pi$.

\textbf{Novelty of states.} In RL, data such as states and rewards are derived from agent-environment interactions. Frequently encountered states, often near the starting point, are well-fitted by neural networks, leading to accurate value estimations. Conversely, actions sampled with low probabilities may result in infrequent or unseen states, causing inaccurate estimations. Thus, in RL, the frequency of state occurrences serves as a measure of novelty, with less frequent states considered more novel.

Currently, various methods exist to quantify state novelty. A common approach is to count the occurrences of observed states or use probability distributions for pseudo-counts, as exemplified by methods such as CTS \cite{bellemare2016unifying}, PixelCNN \cite{ostrovski2017count}, and Successor Counts \cite{machado2020count}. Another prevalent method involves using dynamical models to predict the next state based on the current state and action \cite{achiam2017surprise,pathak2017curiosity}, where prediction accuracy correlates with the frequency of state occurrences, and the prediction error serves as a measure of novelty. An alternative approach involves network distillation, exemplified by RND \cite{burda2018exploration}, where a prediction network distills a random network, and the discrepancy between the random target network and the prediction network indicates novelty, with smaller discrepancies reflecting higher state occurrence frequencies. In our paper, we adopt this third method, utilizing RND to evaluate state novelty.

\begin{figure}[h]
    \centering
    \includegraphics[width=0.5\textwidth]{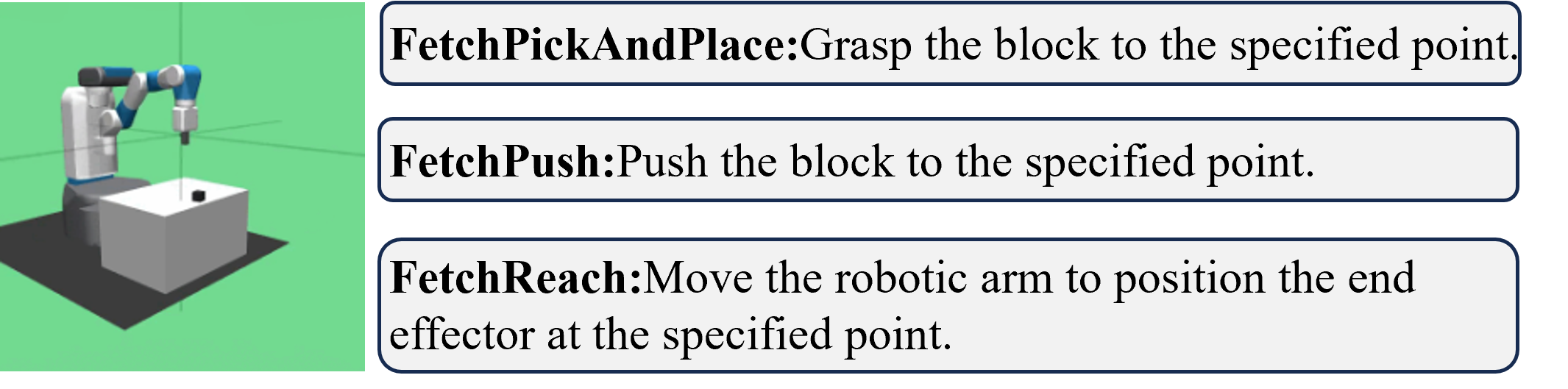}
    \caption{
Visualization of the Fetch environment. The agent needs to control a robotic arm to complete tasks such as grasping and pushing objects to specified locations.}
    \label{fig2}
\end{figure}

\section{Method}
\subsection{Enhancing Convergence Rate Through Sample Reuse}
In reinforcement learning, data is obtained through interactions between the agent and the environment: the agent performs actions, and the environment returns the next state and reward. Unlike supervised learning, which directly retrieves data from a dataset, this interaction process is often time-consuming and generally requires more time than the actual updating of network parameters.

In on-policy reinforcement learning algorithms like PPO, it is vital to maintain similarity between the sampling and current policies, as collected data is discarded after use. Off-policy methods, on the other hand, randomly sample from a replay buffer for updates, requiring further interactions with the environment. This inefficiency leads to underutilization of historical experiences, causing the agent to operate on a sub-optimally trained policy and resulting in lower-quality data collection and higher interaction costs.

To address this issue, we employ a sample reuse method, allowing repeated updates of network parameters using samples obtained from environment interactions or the replay buffer. This approach enhances the agent's utilization of historical data, enabling it to refine its policy before further interactions with the environment. We conducted experiments on the `Fetch-PickAndPlace' task, comparing the effects of updating the network parameters once, twice, three times, and five times. Notably, reusing samples only increases the time required for network updates and does not increase the time spent interacting with the environment. Therefore, the overall run time does not increase significantly. The experimental results show that reusing data samples results in a higher task completion rate with the same number of steps, demonstrating the importance of sample reuse.

\begin{figure*}[t]
    \centering
    \includegraphics[width=0.95\linewidth]{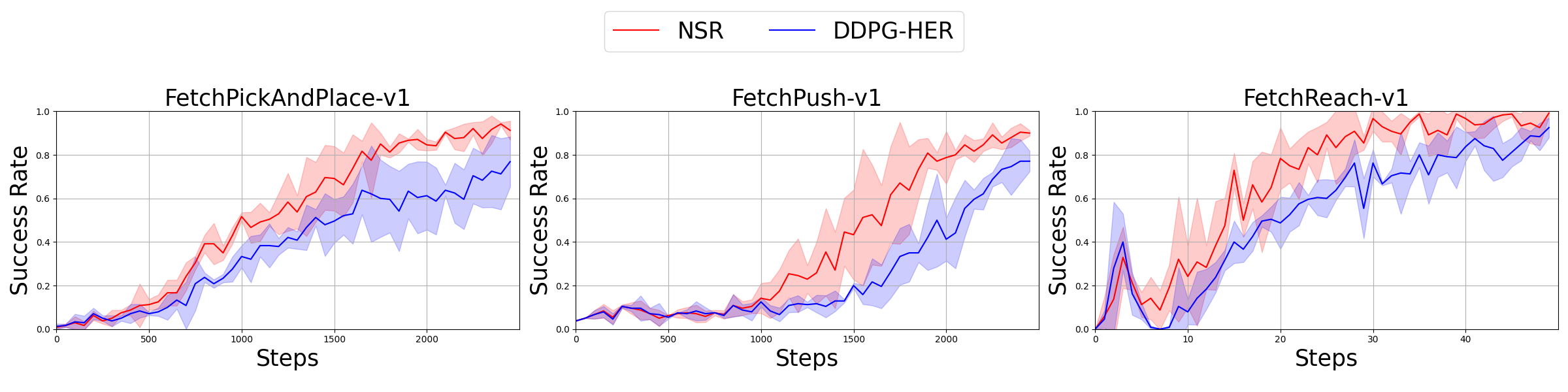}
    \caption{Results for manipulation tasks are averaged over five runs, with the shaded area representing the standard deviation.}
    \label{fig3}
\end{figure*}

\subsection{Adaptive Sample Utilization Based on State Novelty}
The previous section experimentally validated the effectiveness of sample reuse. During training, this approach allows the agent to fully utilize historical data, improving task completion rates and reward growth after learning an effective strategy. However, for frequently encountered states, such as those near the starting point, the replay buffer often contains many similar samples. In these cases, critic values are well-established, making further reuse unnecessary. Conversely, for infrequently visited states, value estimations and strategies tend to be inaccurate due to their rarity, requiring repeated use of these samples to refine the agent's strategy.

To address this issue, we incorporated state novelty. When a state demonstrates higher novelty, we are inclined to update that state more frequently. To quantify this novelty, we employed the curiosity-driven Random Network Distillation (RND) method from reinforcement learning exploration. By evaluating the discrepancy between the prediction and target networks, we determine whether the current state is novel, which in turn informs the need for additional updates.

In detail, the RND method involves a target network \( f_{\bar\theta} \) with randomly given fixed parameters that produces a multi-dimensional random output when given an input, and a prediction network \( f_{\theta} \) that is randomly initialized and needs to be updated. Each time a state \( s \) is encountered, the target network and the prediction network output \( f_{\bar\theta}(s) \) and \( f_{\theta}(s) \), respectively. MSE loss is used to force the prediction network to distill and align with the target network's output. Since the target network's output is random and cannot be predicted from other states, the MSE loss will only decrease when the current state is visited multiple times, according to the characteristics of neural network fitting. Conversely, for states that are rarely or never seen, this MSE loss will be large. The RND method uses this MSE loss as a measure of state novelty and applies it to address agent exploration and sparse reward scenarios.

Unlike the approach of using novelty for agent exploration to address sparse rewards, we focus on sample utilization in typical reinforcement learning scenarios with environmental rewards. High-novelty states are rarely sampled, resulting in imprecise value function estimates that require more updates to reduce prediction error. By reusing this data for multiple updates, we increase the visit frequency of these infrequent samples, preventing the learning state distribution from becoming imbalanced and biased. We term our method Novelty-guided Sample Reuse (NSR), which provides adaptive sample utilization based on state novelty.

\subsection{Combine DDPG with NSR}
In this paper, we combine the NSR method with the DDPG algorithm to address the problem of continuous control. When the input state $s_t$ is given, we use the RND method to calculate the novelty $N^\theta_\text{mse}(s_t)=\|f_{\theta}(s_t)-f_{\bar\theta}(s_t)\|^2$ of the state and then perform normalization to standardize, i.e., $N_
\theta(s_t)=\frac{N_\text{mse}(s_t)-N_\text{mse}(s_t).\text{mean}()}{N_\text{mse}(s_t).\text{std}()}$. Since standard normal normalization can result in values less than 1 or negative values, to ensure that all samples are learned and only the novel states are updated additionally, we set the minimum value of $N(s_t)$ to 1. To avoid introducing significant additional overhead, we cap the maximum value of $N(s_t)$ at 3, meaning a state will not be updated more than three times.

In our experiments, we discovered that by multiplying the original loss with a coefficient, we achieved similar effects as multiple updates. This approach eliminates the need for additional iterations and reduces computational overhead. For more detailed results, please refer to the experiment section. Our NSR method multiplies the calculated $N(s_t)$ with both the actor and critic loss functions, resulting in the modified loss as follows:
\begin{equation}
    \small
    \mathcal{L}_\text{critic}(\varphi)=\|N(s_t)[Q_\varphi(s_t,\mu_\phi(s_{t}))-r_t-\gamma Q_\varphi(s_{t+1},\mu_\phi(s_{t+1}))]\|^2
\end{equation}

\begin{equation}\small\mathcal{L}_\text{actor}(\phi)=-N(s_t)Q_\varphi(s_t,\mu_\phi(s_{t}))
\end{equation}

\begin{equation}\small
    \mathcal{L}_\text{RND}(\theta)=\|f_\theta(s_t)-f_{\bar\theta}(s_t)\|^2.
\end{equation}

By using this method, we can leverage the novelty of the state within the DDPG algorithm to assign different update weights to different samples, thereby approximately achieving the effect of updating each sample a different number of times. The overview of our method is illustrated in Figure \ref{overview}.

\begin{figure*}[t]
	\centering
	\begin{minipage}[c]{0.95\textwidth}
		\centering
		\includegraphics[width=\textwidth]{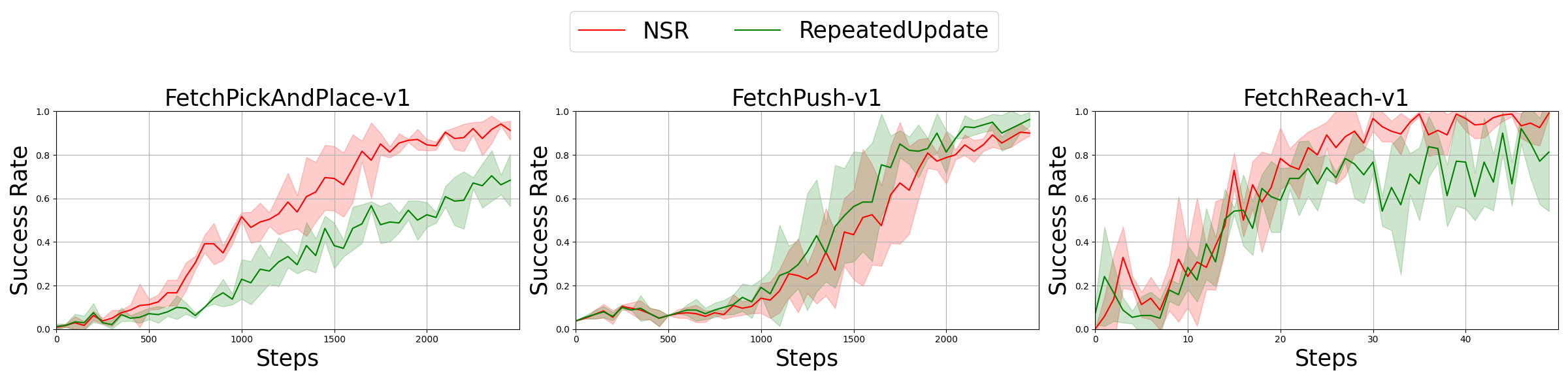}
		\subcaption{}
		\label{fig_4_1}
	\end{minipage} \\
	\begin{minipage}[c]{0.95\textwidth}
		\centering
		\includegraphics[width=\textwidth]{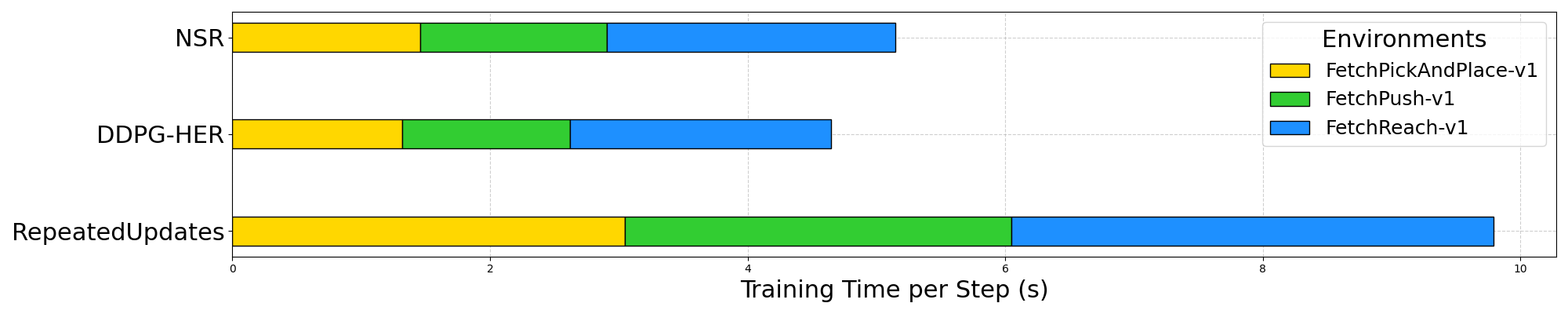}
		\subcaption{}
		\label{fig_4_2}
	\end{minipage}
	\caption{(a) The performance of NSR and DDPG-HER is generally equivalent, in the FetchPush environment, repeated updates yield slightly better results at a significantly higher time cost. (b) During training, repeated updates require considerably more time than DDPG-HER, while NSR incurs only minimal additional overhead.}
	\label{fig_4}
\end{figure*}

\begin{figure*}[t]
    \centering
    \includegraphics[width=0.95\linewidth]{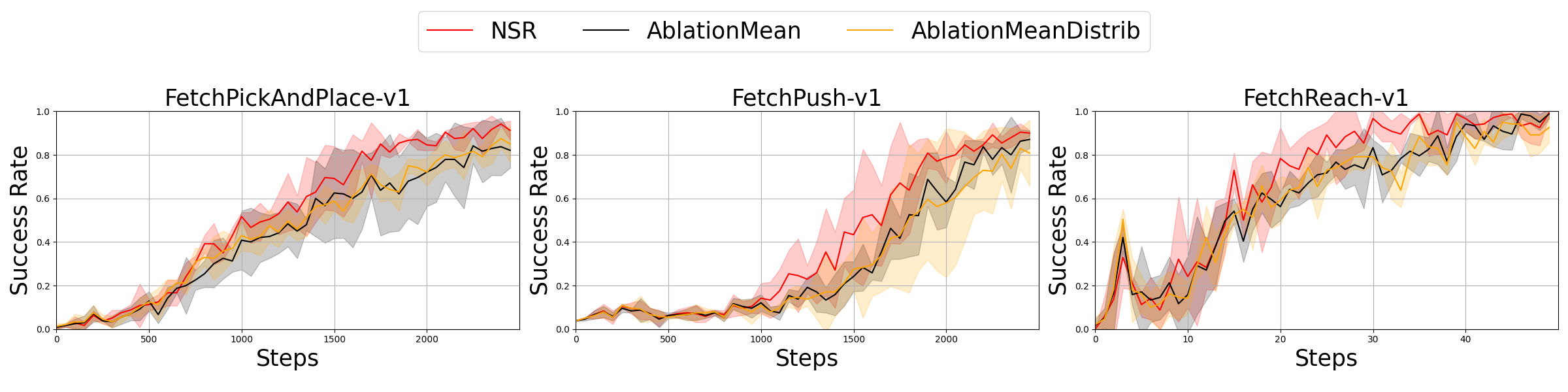}
    \caption{NSR outperforms mean weights and random weights in the ablation study, highlighting its effectiveness.}
    \label{fig5}
\end{figure*}

\section{Experiment}
In this section, we will demonstrate the effectiveness of NSR method in continuous control environments through experiments. We have chosen the widely used and effective DDPG-HER \cite{2017Hindsight} algorithm as our baseline for continuous control and integrated our method with it. To validate our algorithm, we selected the Fetch Manipulation robotic arm environment \cite{plappert2018multi} from gym robotics as our testing environment.

\subsection{Setup}
We use three environments from Fetch: Reach, Push, and PickAndPlace as our validation environments. Examples of robotic arm and environments are shown in Figure \ref{fig2}. In these environments, the robotic arm performs tasks such as touching, pushing a block to a designated location, and grasping a target block, while tracking the success rate of these actions. All our experimental environments used the `v1' version, with the baseline algorithm configured with the following parameters: batch\_size = 256, actor\_lr = 1e-3, critic\_lr = 1e-3, and gamma = 0.98. We set the maximum training steps to 2500 and the maximum number of interaction steps with the environment to 50. The network architecture consists of 4 linear layers, with ReLU as the activation function in the hidden layers, each with a dimension of 256. For more detailed settings, you can refer to the code link we provided in the abstract.

\subsection{Result}
The efficiency of DDPG-HER and NSR in managing complex robotic manipulations is highlighted in our evaluation of Fetch manipulation tasks. As shown in Figure \ref{fig3}, in all tasks, NSR achieves a higher success rate with the same number of training steps, demonstrating the effectiveness of NSR.

NSR evaluates state novelty and assigns importance levels accordingly, giving greater weights to states with greater novelty. This approach allows NSR to achieve comparable results while avoiding the additional computational cost of repeated updates. The specific results are shown in Figure \ref{fig_4}.

\subsection{Ablation Study}
To determine whether the improved training efficiency was primarily attributed to the reuse of novel samples or the enhanced learning rate resulting from the introduction of state weights, we conducted the following experiments: Firstly, to ensure a fair comparison of methods in terms of learning rate, we computed the mean of all state weights at each step and evenly distributed this mean weight among all states. Then, In order to validate the significance of novelty-guided weight assignment, we utilized the mean and variance weights to create normal distributions for assigning state weights. As shown in Figure \ref{fig5}, NSR demonstrates superior performance compared to the other two weight assignment methods, thus highlighting the critical role of novelty.

\section{Conclusion}
In this paper, we introduce a novel approach called NSR that assesses the necessity for additional uses of a state based on its novelty, optimizing the update frequency for each sample to enhance sample efficiency. Our method maximizes the utilization of samples from the environment, engaging in interactions only after thorough learning, which increases the ratio of sample utilization to environmental interaction. By using weights rather than update counts, we achieve varied update frequencies among samples with minimal computational overhead. Experimental results demonstrate that our approach performs comparably to repeated updates while significantly surpassing baseline algorithms across multiple tasks in the Fetch robotic continuous control simulation environment, thereby accelerating task convergence. Ablation studies further validate the effectiveness of our method.

\section{Acknowledgements}
This work was supported by the Natural Science Foundation of China under Grant 92248304 and Shenzhen Science Fund for Distinguished
Young Scholars under Grant RCJC20210706091946001.

%
%
%
\bibliographystyle{splncs04}
\bibliography{ref}
%




\end{document}